\documentclass{article}
\usepackage{ijcai22}

\usepackage{mathrsfs}
\usepackage{times}
\usepackage{soul}
\usepackage{url}
\usepackage[hidelinks]{hyperref}
\usepackage[utf8]{inputenc}
\usepackage[small]{caption}
\usepackage{graphicx}
\usepackage{amsmath}
\usepackage{amsthm}
\usepackage{booktabs}
\usepackage{algorithm}
\usepackage{algorithmic}
\usepackage{tablefootnote}
\usepackage{amsfonts}

\title{Learning Neural Hamiltonian Dynamics: A Methodological Overview}

\author{
Zhijie Chen$^{1\#}$\and
Mingquan Feng$^{1\#}$\and
Junchi Yan$^1$\footnote{The first two authors contributed equally to this paper. Correspondence author is Junchi Yan.}\and 
Hongyuan Zha$^{2}$
\affiliations
$^1$Department of CSE, and MoE Key Lab of Artificial Intelligence, 
Shanghai Jiao Tong University\\
$^2$ Chinese University of Hong Kong, Shenzhen
\emails  
\{chen-zhijie,fengmingquan,yanjunchi\}@sjtu.edu.cn
\quad  zhahy@cuhk.edu.cn
}

\begin{document}

\maketitle

\begin{abstract}
  The past few years have witnessed an increased interest in learning Hamiltonian dynamics in deep learning frameworks. As an inductive bias based on physical laws, Hamiltonian dynamics endow neural networks with accurate long-term prediction, interpretability, and data-efficient learning. However, Hamiltonian dynamics also bring energy conservation or dissipation assumptions on the input data and additional computational overhead.  In this paper, we systematically survey recently proposed Hamiltonian neural network models, with a special emphasis on methodologies. In general, we discuss the major contributions of these models, and compare them in four overlapping directions: 1) generalized Hamiltonian system; 2) symplectic integration, 3) generalized input form, and 4) extended problem settings. We also provide an outlook of the fundamental challenges and emerging opportunities in this area.
\end{abstract}

\section{Introduction}
Generating data from laws is physics; Inferring laws from data is physics-informed learning~\cite{raissi2019physics}. These two classes of works representing \textit{forward problems} and \textit{inverse problems} are at the center of the  research. The past few years have witnessed ramping-up interests in integrating physics-informed priors into deep neural network architectures, which give new insights into both directions. These architectures are often highly bound with a specific class of differential equations, which are typically described as $\mathbf{\dot z} = f( \mathbf{z}, t)$. With the advent of NeuralODE~\cite{chen2018neural}, a general parametric framework for the inverse problem is brought to a large audience. It carries out forward integration and cleverly updates the parameters by the adjoint system. However, since NeuralODE is an end-to-end blackbox model that conceals underlying dynamics beneath the integrator, it is often not expected to model the exact trajectory in a physical system. On the other hand, an orthogonal line of work focuses on the forward problem. This includes the Deep Ritz Method~\cite{weinan2018deep} that learns trial functions for the Ritz method in solving PDEs and Neural Fourier Operators~\cite{kovachki2021neural,li2020fourier} which typically learn a functional map from the parameter space to the solution space in a data-driven manner. Physics-Informed Neural Network (PINN)~\cite{han2018solving} solves both the data-driven differential equation solution and the discovery problem by utilizing the universal approximation capability of neural networks. While these models are designed for generic differential equations, it is also promising to further impose inductive biases if the underlying dynamics have distinguishing characteristics.

As one of the most profound insights in physics, Noether tells that all of the physical laws can be expressed as conserved quantities. For example, the time translation invariance implies the conservation of energy in a physical system. This observation further gives rise to the formulation of Hamiltonian system, which has nowadays become a fundamental physical system widely applied to various cases.
 
Hamiltonian-based neural networks serve as a framework that considers both physical dynamics learning and prediction, with applicability in both the forward problem and inverse problem. It is known that even a simple formulation of the Hamiltonian system can suffer from divergence induced by accumulative discretization and truncation error. Hence a question naturally arises:

\textit{Q1: Given the Hamiltonian and an initial value, how to predict the trajectory with high fidelity?} 

On the other side, The Hamiltonian (including kinetic energy and potential energy) is of particular interest for a deeper understanding of the underlying physical system, which induces our second question:

\textit{Q2: Given a trajectory of motions (possibly with noise), how to infer the Hamiltonian by the state of the iterate?}

To answer the above two questions, endeavors have been made on the track of incorporating Hamiltonian systems into deep neural networks. This practice is advantageous especially in the following aspects:
\begin{itemize}
    \item \textit{Low Memory Requirements.} Incorporating inductive biases in the deep models reduces the function searching space, and results in significantly fewer trainable parameters. It is reported \cite{dipietro2020sparse} that under adequate settings, the training parameters are reduced from billions to thousands compared with a blackbox prediction model (with comparable performance).
    \item \textit{Interpretability.} We know only a small set of laws are governing various physical systems. Since the Hamiltonian-based neural networks strictly observe the laws, they can give hints to explain the natural phenomena by providing interpretable counterparts in physical laws.
    \item \textit{High predictive accuracy.} Hamilton system learning adopts an innovative idea of learning the Hamiltonian that generates the vector field, instead of learning the vector field itself, whereby these models eliminate the approximation error of learning the vector field, which is presumably inefficient and problematic in high-dimensional spaces.
    \item \textit{Improved Data Efficiency and Generalization.}  Generalizations in neural networks can be improved by leveraging underlying physics for designing computation graphs. It helps extrapolation and is not sensitive to out-of-sample behavior, leading to few-shot learning scenarios~\cite{sanchez2019hamiltonian}.
    \item \textit{Facilitating Downstream Usage.} Learning a physically-consistent model informed by the physical environment can be adopted as a building block of model-based controllers, which is prevalent in the context of reinforcement learning~\cite{polydoros2017survey}. In particular, one can leverage the kinetic and potential energy learned in a Hamiltonian system to synthesize appropriate control strategies, such as the method of controlled Lagrangian~\cite{bloch2001controlled} and interconnection  damping assignment~\cite{ortega2002interconnection}, which may reshape the current closed-loop  energy landscape.
    \item \textit{Wide real applications.} Hamiltonian-based learning is exceptionally suitable for physical systems that obey the Hamiltonian dynamics, with possible extension to nonlinear and chaotic systems like a double pendulum and n-body systems~\cite{chen2019symplectic,choudhary2020physics}. It also shows power in various domains including Schrödinger Hamiltonian for quantum mechanics~\cite{valenti2019hamiltonian,dutt2021active,rupp2012fast}, Hamilton-Jacobi Equations for optimal control~\cite{yang2017hamiltonian} , numerical optimization for minimax games~\cite{letcher2019differentiable} and ordinary differential equations~\cite{matsubara2021symplectic}.
\end{itemize}

This paper gives a systematic review of the recent advances of Hamiltonian-based learning and symplecticity-preserved integration. Our main contributions are: 

1) We give a methodological view of learning paradigms for Hamiltonian dynamics. We focus on the methods that are adopted under different settings. It includes works targeting at both classical and generalized formulations of Hamiltonian dynamics. And we also explore the models that admit either canonical or fancy input formats. 

2) We pay attention to the integration methods which also play an important role in predicting the Hamiltonian dynamics. Specifically, the usage of different kinds of symplectic integrators and the way they are intertwined with neural architecture is of our primary interest.

\textbf{Related work} 
There are a plethora of surveys about deep learning with physical priors~\cite{wang2021physics,willard2020integrating}. These surveys study the broader area of applications, e.g. fluid dynamics and molecular dynamics, whereas our main focus is Hamiltonian dynamics.

There is also some effort in summarizing recent advances relevant to learning  Hamiltonian dynamics. In particular, \cite{zhong2021benchmarking} is perhaps the most relevant to our work, as it benchmarks ten energy-conserving neural network models, including HNN~\cite{greydanus2019hamiltonian}, SymODEN~\cite{zhong2019symplectic} and CHNN~\cite{celledoni2022learning}, and also Lagrangian models like Deep Lagrange Network (DeLaN)~\cite{lutter2018deep}. It compares the performance of these models on four synthetic physics systems, and provides theoretical derivation of the constrained HNN learning. However, our work focuses on Hamiltonian Dynamics, and gives a much wider coverage in this promising area and also in-depth analysis to the existing works.

\section{Preliminaries}

\subsection{Hamiltonian Dynamics}
We start by considering a standard form of Hamiltonian system which is essentially a reformulation of Newtonian dynamics and yet provides fundamental insights into the laws of mechanics. More specifically, a $2n$-dimensional Hamiltonian system traces the temporal evolution of the system states in the phase space, i.e. the product space of generalized coordinates $\mathbf{q} = (q_1, q_2,...,q_n) \in \mathcal{Q} \subseteq \mathbb{R}^n$ and generalized momenta $\mathbf{p} = (p_1, p_2,...,p_n) \mathcal{P} \subseteq \mathbb{R}^n$. Over time, the state points move in the vector field of the Hamiltonian system forms a unique motion trajectory. Note that the incorporation of generalized coordinates is formed from any set of variables that describe the complete state of the system and thus can be carefully designed to implicitly ensure some of the real-world constraints to hold automatically~\cite{finzi2020simplifying}. 

This time evolution of a dynamical system is described by a scalar function termed as Hamiltonian $H(\mathbf{q}, \mathbf{p})$, which takes the generalized coordinates and momenta as input. In the context of classical mechanics, the Hamiltonian is equivalent to the total energy of a physical system and typically takes the form of
\begin{equation}
    H(\mathbf{q}, \mathbf{p}) = \frac{1}{2} \mathbf{p}^\top \mathbf{M}^{-1}(\mathbf{q})\mathbf{p} + V(\mathbf{q}),
\end{equation}
where the mass matrix $\mathbf{M}(q)$ is symmetric positive definite (SPD). The term $\frac{1}{2} \mathbf{p}^\top \mathbf{M}^{-1}(\mathbf{q})\mathbf{p}$ represents the kinetic energy and $V(\mathbf{q})$ represents the potential energy of the system. Correspondingly, the time-evolution of the system is governed by
\begin{equation}
    \mathbf{\dot q} = \frac{\partial H}{\partial \mathbf{p}} ~~~~~~~~~\mathbf{\dot p} = -\frac{\partial H}{\partial \mathbf{q}},
    \label{eq:hamilton}
\end{equation}
The direction of the vector field defined by Eq.~\ref{eq:hamilton} is often called the \textit{symplectic gradient} of the Hamiltonian system. A straightforward implication of the evolution function~\ref{eq:hamilton} is 
\begin{equation}
    \dot H = \left(\frac{\partial H}{\partial \mathbf{q}} \right)^\top \mathbf{\dot q} + \left(\frac{\partial H}{\partial \mathbf{p}} \right)^\top  \mathbf{\dot p} \equiv 0,
\end{equation}
implying that the total energy (Hamiltonian) is conserved along the motion trajectory defined by the system.

A Hamiltonian system is called separable if the Hamiltonian can be separated into additive terms, each of which is dependent on either generalized coordinates or generalized momenta, i.e. $H(\mathbf{q},\mathbf{p}) = T(\mathbf{q}) + V(\mathbf{p})$, where $T$ and $V$ are arbitrary functions.

\subsection{Symplectic Transformation}
The energy-conserving property of time evolution of Hamiltonian dynamics is a manifestation of \textit{symplectomorphism}, which essentially represents a
transformation of phase space that is volume-preserving. Define an orthogonal, skew-symmetric real matrix $J = \begin{bmatrix} 0 & \mathbb{I}_n \\ \mathbb{I}_n & 0
\end{bmatrix}$, where $\mathbb{I}_n$ is an $n$-by-$n$ identity matrix. A differentiable map $g: U \rightarrow \mathbb{R}^{2n}$ (where $U \subset \mathbb{R}^{2n}$) is called \textit{symplectic} if the Jacobian matrix $\frac{\partial g}{\partial x}$ satisfies $\left(\frac{\partial g}{\partial x} \right)^\top J \left(\frac{\partial g}{\partial x}\right) = J$. More intuitively, we can define a $2n$-dim parallelogram volume function $\omega(\xi, \eta) = \xi^\top J \eta$, for arbitrary vectors $\xi, \eta \in \mathbb{R}^{2n}$. With symplectic mapping carried out to $\xi$ and $\eta$ , the volume of the parallelogram is preserved, i.e. $\omega(\frac{\partial g}{\partial x} \xi, \frac{\partial g}{\partial x}\eta) = \omega(\xi, \eta)$. 

As early as in 1899, Poincaré pointed out in his seminal work~\cite{poincare1899methodes} that the flow induced by a Hamiltonian system is everywhere symplectic as long as the Hamiltonian $H(q,p)$ is twice continuously differentiable on its domain. In spite of being a characteristic property for Hamiltonian system, discrete symplectic mappings do not guarantee a corresponding global Hamiltonian system. One instance is the standard map in accelerator physics \cite{chen2021data}. Investigations in symplectic mappings hence step further beyond classical Hamiltonian dynamics.

Nonetheless, the sufficient condition of a symplectic mapping can be characterized by a differentiable scalar-valued function, known as \textit{generating function} $F(q,P)$. In 1980, Goldstein pointed out that the mapping $(p,q) \rightarrow (P,Q)$ implicitly defined by $p = \frac{\partial F}{\partial q}(q,P)$ and $Q=\frac{\partial F}{\partial P}(q,P)$, is a symplectic mapping. It can be further accommodated to the context of Hamiltonian system by Hamilton-Jacobi PDE: $H(\frac{\partial F}{\partial q}, q, t) + \frac{\partial F}{\partial t} = 0$.

\section{Deep Learning of Hamiltonian Dynamics}
\subsection{Learning Standard Hamiltonian System}
The idea of learning Hamiltonian dynamics by machine learning models dates back to the 1990s \cite{howse1995gradient,seung1997minimax}, recently a ground-breaking study of learning a deep neural network for Hamiltonian, Hamiltonian Neural Networks (HNN) \cite{greydanus2019hamiltonian} emerges. HNN is a neural network that models the Hamiltonian $H$ and defines the
dynamics following the Hamiltonian mechanics, thereby ensuring the energy conservation law. 

\begin{figure}[tb!]
    \centering
    \includegraphics[width=0.49\textwidth]{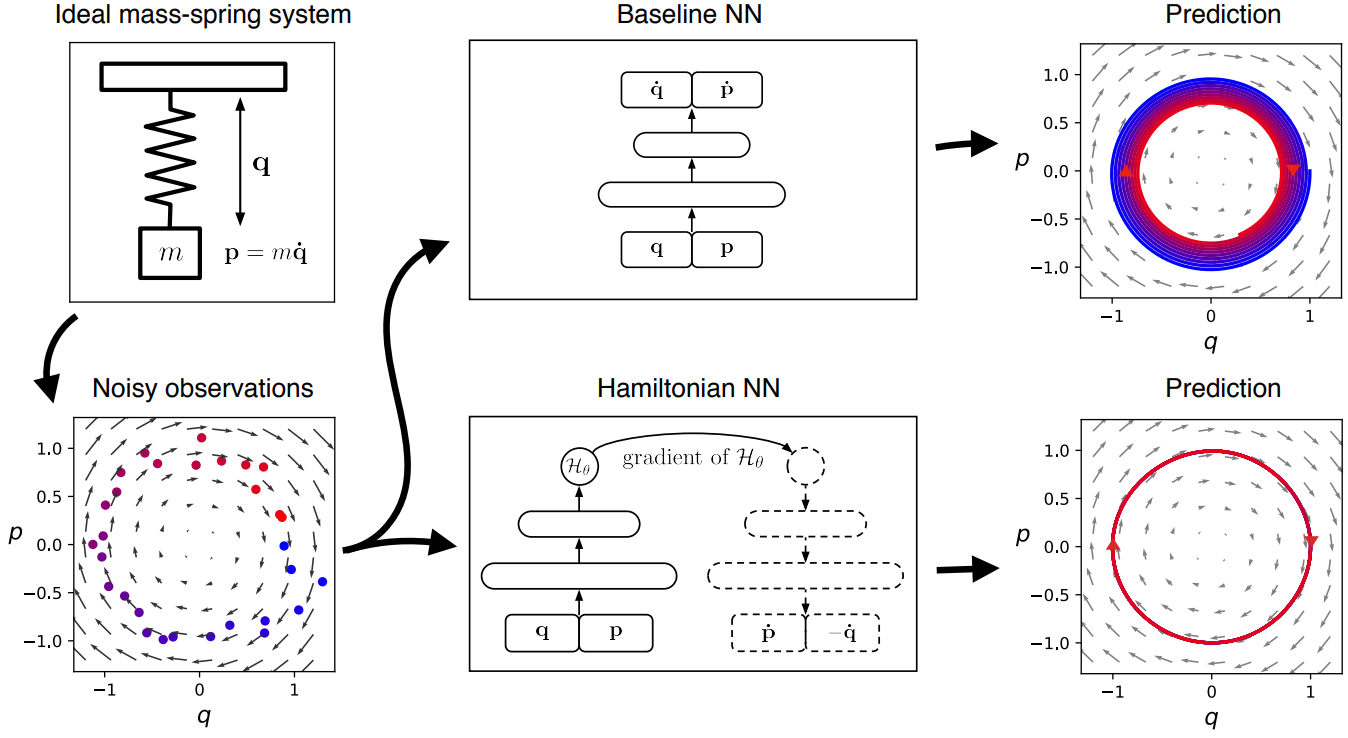}
    \vspace{-5pt}
    \caption{Comparison between baseline NN and Hamiltonian NN (HNN). Variable $q,p$ correspond to position and momentum coordinates, and $\dot{\mathrm{p}} ,\dot{\mathrm{q}}$ denote their derivatives with respect to time. The baseline NN wrongly outputs an inner spiral trajectory due to model errors. In comparison, HNN learns an accurate circle trajectory, as it exactly conserves a quantity analogous to the  total energy. (credit to \protect\cite{greydanus2019hamiltonian})}
    \label{fig:hnn}
\end{figure}

In the forward pass, HNN consumes a set of coordinates $\mathbf{p},\mathbf{q}$ and
outputs a single scalar “energy-like” value $\mathcal{H}_{\theta}$, where $\theta$ denotes trainable parameters. Then, before computing the loss, HNN takes a partial derivative of the output to the input coordinates. Then HNN computes and optimizes the following L2 loss (see Figure \ref{fig:hnn} for illustration):
$$
\mathcal{L}_{H N N}=\left\|\frac{\partial \mathcal{H}_{\theta}}{\partial \mathbf{p}}-\frac{\partial \mathbf{q}}{\partial t}\right\|_{2}+\left\|\frac{\partial \mathcal{H}_{\theta}}{\partial \mathbf{q}}+\frac{\partial \mathbf{p}}{\partial t}\right\|_{2}
$$

Experiments in physical systems (mass-spring, pendulum, and 2-body problem) show that compared with baseline neural networks, HNN trains more quickly and generalizes better, since it learns to conserve the Hamiltonian.

After adding an autoencoder module, HNN is also able to deal with image input, e.g. the Pendulum-v0 environment in OpenAI Gym \cite{brockman2016openai}. The input images are encoded to latent vector $\left(\mathbf{z}_{\mathbf{q}}, \mathbf{z}_{\mathbf{p}}\right)$, and the loss becomes 3 terms: 1)$\mathcal{L}_{H N N}$, 2)autoencoder loss $\left(L_{2}\right.$ loss over pixels $)$ and 3) $\mathcal{L}_{C C}=\left\|\mathbf{z}_{\mathbf{p}}^{t}-\left(\mathbf{z}_{\mathbf{q}}^{t}-z_{\mathbf{q}}^{t+1}\right)\right\|_{2}$. The last loss term enforces the latent vector $\left(\mathbf{z}_{\mathbf{q}}, \mathbf{z}_{\mathbf{p}}\right)$ to have roughly same properties as canonical coordinates $(\mathbf{q}, \mathbf{p})$. 

The successful application of HNN intrigues many subsequent works and ideas, as we will discuss in the following sections.

\subsection{Learning Generalized Hamiltonian System}
\label{sec:general-system}
The standard Hamiltonian system is a model for conservative systems, in the sense that an object moving in the direction of the symplectic gradient $S_{\mathcal{H}}=\left(\frac{\partial \mathcal{H}}{\partial \mathbf{p}},-\frac{\partial \mathcal{H}}{\partial \mathbf{q}}\right)$ keeps the Hamiltonian as time-invariant. However, real physical systems are often nonconservative systems, since they involve dissipative dynamics due to friction, external
forces, or measurement noise.

\begin{figure}[tb!]
    \centering
    \includegraphics[width=0.48\textwidth]{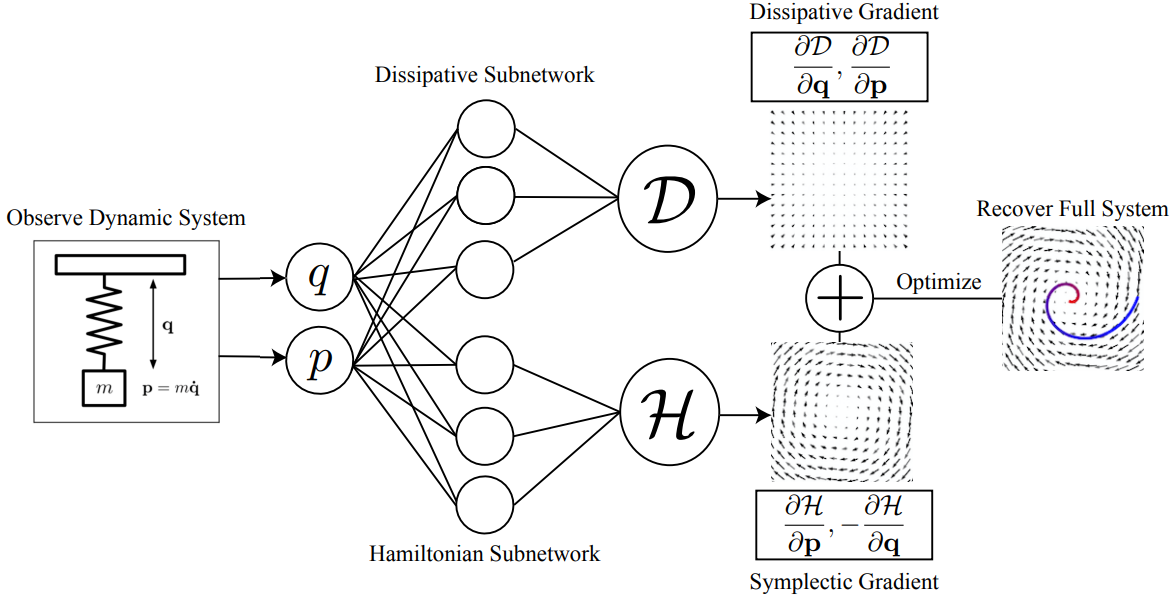}
    \vspace{-5pt}
    \caption{Model architecture of Dissipative Hamiltonian Neural Networks (DHNN), a typical model for the generalized Hamiltonian system. The figure visualized a dynamic system of a damped spring (e.g. a spring with friction), where the system is decomposed into a dissipative system (upper part of the figure) and a conservative system (lower part of the figure). The former is visualized as a dissipative vector field, denoting the energy loss caused by friction. 
    (credit to \protect\cite{greydanus2022dissipative})}
    \label{fig:dhnn}
\end{figure}

To predict dynamics in dissipative environment, Dissipative Hamiltonian Neural Networks (DHNN)  \cite{greydanus2022dissipative} generalizes HNN by adding a  Rayleigh dissipative function $\mathcal{D}(\mathbf{q}, \mathbf{p})$. The system now becomes:
$$
\frac{d \mathbf{q}}{d t} = \frac{\partial \mathcal{H}}{\partial \mathbf{p}}+\frac{\partial \mathcal{D}}{\partial \mathbf{q}}, \quad
\frac{d \mathbf{p}}{d t} = -\frac{\partial \mathcal{H}}{\partial \mathbf{q}}+\frac{\partial \mathcal{D}}{\partial \mathbf{p}}
$$
The functions $\mathcal{H}$ and $\mathcal{D}$ are both parameterized as neural networks, and the parameters are optimized via L2 loss of the above equations. The model architecture is shown in Figure \ref{fig:dhnn}.

Theoretically, DHNN performs an implicit Helmholtz decomposition \cite{helmholtz1858integrale}, which states that any smooth vector field $V$
can be expressed as a sum of its irrotational component $V_{irr}$ (approximated by $\mathcal{D}$)
and its rotational component $V_{rot}$ (approximated by $\mathcal{H}$). 

The Hamiltonian system can be further generalized, as proposed in \cite{howse1995gradient}. In essence, the Hamiltonian system aims at modeling the time-invariant/variant value, which can be defined by the following system:
$$
\frac{d\mathbf{x}}{dt}=(\mathbf{P}(\mathbf{x})+\mathbf{Q}(\mathbf{x})) \nabla \mathcal{H}(\mathbf{x}),
$$
where $\mathbf{x} \in \mathbb{R}^{n}$ is the state of dynamics, $\mathbf{P}: \mathbb{R}^{n} \rightarrow \mathbb{R}^{n \times n}$ is a skew-symmetric matrix depending on $\mathbf{x}$ (i.e. $\mathbf{P}(\mathbf{x})^T= - \mathbf{P}(\mathbf{x})$),  $\mathbf{Q}: \mathbb{R}^{n} \rightarrow \mathbb{R}^{n \times n}$ is a symmetric matrix depending on $\mathbf{x}$. It is easy to verify that the first part of above system, i.e. $\frac{d\mathbf{x}}{dt}=\mathbf{P}(\mathbf{x})\nabla \mathcal{H}(\mathbf{x})$, denotes the time-invariant $\mathcal{H}$, while the second part of the system, i.e. $\frac{d\mathbf{x}}{dt}=\mathbf{Q}(\mathbf{x})\nabla \mathcal{H}(\mathbf{x})$, models the time-variant $\mathcal{H}$. Note that we can
recover the standard Hamiltonian decomposition by setting $\mathbf{x}=\left(\begin{array}{l}\mathbf{q} \\ \mathbf{p} \end{array}\right)$ and :
$$
\mathbf{P}(\mathbf{x})=\left[\begin{array}{cc}
\mathbf{0} & \mathbf{1} \\
-\mathbf{1} & \mathbf{0}
\end{array}\right] \quad \& \quad \mathbf{Q}(\mathbf{x})=\mathbf{0}.
$$

Such a generalized form Hamiltonian system has larger model capacity, since the state $\mathbf{x}$ can be odd-dimensional, and coefficient matrices $\mathbf{P},\mathbf{Q}$ can be state-dependent. However, the improvement of capacity also leads to the risk of overfitting or instability, thus this generalized model is usually applied with additional regularization. 

Generalized HNN (GHNN) \cite{course2020weak} enforces global stability by parameterizing $\mathbf{Q}(\mathbf{x}) \nabla \mathcal{H}(\mathbf{x})=\nabla^{2} \mathcal{N}_{v}(\mathbf{x}) \nabla \mathcal{N}_{\mathcal{H}}(\mathbf{x})$, where $\mathcal{N}_{v}$ is chosen to be an input concave neural network \cite{amos2017input}. And $\mathcal{N}_{\mathcal{H}}(\mathbf{x})=\operatorname{ReHU}(\mathcal{N}(\mathbf{x})-\mathcal{N}(\mathbf{0}))+\epsilon \mathbf{x}^{\top} \mathbf{x}$, where $ReHU$ is the rectified Huber unit. Thus GHNN can be guaranteed to be globally asymptotically stable, i.e. always converge to $\mathbf{x}=\mathbf{0}$ in finite time. To accelerate integration, GHNN multiplies ODE by test functions, and the result can be efficiently estimated using standard quadrature techniques.

GENERIC \cite{lee2021machine} also learns a general hamiltonian system. To avoid the degeneracy conditions, e.g. the coefficient matrices $\mathbf{P},\mathbf{Q}$ are both zero matrices, it enforces the degeneracy condition
by a carefully designed soft penalty. 

Neural Symplectic Form (NSF)~\cite{chen2021neural} takes a step further by generalizing hamiltonian system to a symplectic 2-form. Suppose phase space is $\mathcal{M}=\mathbb{R}^{2 N} .$ A differential 2 -form $\omega$ on $\mathcal{M}$ is a skew-symmetric bilinear function $$
\omega_{u}\left(v_{1}, v_{2}\right)=v_{1}^{\top} W_{u} v_{2}, \quad \text { for all } v_{1}, v_{2} \in \mathbb{R}^{2 N},
$$ 
where $W_{u}$ is a skew-symmetric matrix. And a symplectic 2-form is a differential 2-form that is nondegenerated and closed. Based on this, the Hamiltonian system is defined as:
$$
\frac{\mathrm{d} u}{\mathrm{~d} t}=X_{H}, \omega\left(X_{H}, v\right)=\mathrm{d} H(v) \quad \text { for all } v \in \mathbb{R}^{2 N},
$$
where $\mathrm{d} H$ is the Fréchet derivative of the Hamiltonian $H$, and $X_{H}$ is a vector field depending on $H$.

To reduce searching space, instead of learning the symplectic 2-form directly, it proposes to learn the differential 1-form of which exterior derivative gives the symplectic 2-form:
$$
\tilde{\omega}=\mathrm{d} \theta_{\mathrm{NN}}, \quad \frac{\mathrm{d} u}{\mathrm{~d} t}=\tilde{X}_{H_{\mathrm{NN}}}, \quad \omega\left(\tilde{X}_{H_{\mathrm{NN}}}, v\right)=\mathrm{d} H_{\mathrm{NN}}(v) 
$$

\begin{table*}[!tb]
\centering
    \resizebox{0.96\linewidth}{!}{
\begin{tabular}{ccccccc}
\toprule
     \textbf{Method}& \multicolumn{1}{c}{\textbf{Hamiltonian Type}} & \multicolumn{1}{c}{\textbf{Loss Form}} & \multicolumn{1}{c}{\textbf{Input Form}} & \multicolumn{1}{c}{\textbf{Separablity Assumption}} & \multicolumn{1}{c}{\textbf{Integration}} \\ \hline
HNN~\cite{greydanus2019hamiltonian}  & Standard  &  Pointwise &  Canonical/Pixel  & No  & Euler  \\ 
DHNN~\cite{greydanus2022dissipative} & Generalized & Pointwise & Canonical & No  &Euler \\
GHNN~\cite{course2020weak} & Generalized & Weak Form & Canonical & No & Euler\\
GENERIC~\cite{lee2021machine}& Generalized & Quadrature & Canonical & No &   Dormand–Prince \\
SRNN~\cite{chen2019symplectic} & Standard   &  Quadrature  &  Canonical/Pixel   &   Yes  &  Leapfrog  \\ 
VIN~\cite{saemundsson2020variational} & Standard & Quadrature &  Canonical/Pixel &No & Velocity Verlet\\
SymODEN~\cite{saemundsson2020variational} & Standard & Quadrature &  Canonical/Angle &No & RK\\
HGN~\cite{toth2019hamiltonian} & Standard & Quadrature &  Canonical/Pixel & Yes & Leapfrog\\
\bottomrule
\end{tabular}
}
\caption{Methodological comparison of Hamiltonian-based Neural Networks. The \textbf{Hamiltonian Type} denotes how Hamiltonian is defined, where the Standard type is only suitable for the conservative system, and General type is for both conservative and dissipative systems. The \textbf{Loss Form} denotes the version of objective governing equations, where Pointwise denotes differential version of Hamiltonian equation, Quadrature denotes integral version, and Weak Form denotes using testing function approximation.   
The \textbf{Input Form} means the acceptable input to the proposed Hamiltonian-based neural networks, where "Canonical" means the phase space $(\mathbf{q}, \mathbf{p})$, and "Pixel" means images/videos. The \textbf{Separability Assumption} means whether the underlying Hamiltonian is needed to be separable. The \textbf{Integration} means which integrator backbone is used to recover the Hamiltonian dynamics from the learned Hamiltonian. Note that we choose the basic integrator in the work (which is a minimum requirement) to represent.}
\label{tab:1}
\end{table*}

\subsection{Symplectic Integration} 
\label{sec:symplectic-integration}
In real-world applications, Hamiltonian systems can often be \emph{stiff}, which means small discretization errors and measurement noise may lead to dramatically diverging trajectories, and thus one has to set time-steps of integration very small to maintain stable numerical solutions. To improve the numerical stability, symplectic integrators, e.g.  the well-known leapfrog method \cite{leimkuhler2004simulating}, are leveraged in HNN models. An integrator of Hamiltonian system is called symplectic if its flow maps are symplectic for short enough time-steps \cite{haier2006geometric}. 

Symplectic Recurrent Neural Network (SRNN) \cite{chen2019symplectic} proposes to integrate partial derivatives of the $\mathcal{H}_{\theta}$ of HNN via the leapfrog integrator, and the loss is back-propagated through the ODE integration over multiple time steps. It assumes that Hamiltonian to be time-invariant and \emph{separable}, i.e. it can be written as $\mathcal{H}(\mathbf{p}, \mathbf{q})=\mathbf{K}(\mathbf{p})+\mathbf{V}(\mathbf{q})$. Thus the hamiltionian system becomes:
$$
\frac{d \mathbf{q}}{d t}= \mathbf{K}^{\prime}(\mathbf{p}), \quad \frac{d \mathbf{p}}{d t}=- \mathbf{V}^{\prime}(\mathbf{q}).
$$
The leapfrog algorithm is then defined as:
$$
\begin{aligned}
\mathbf{p}_{n+1 / 2} &=\mathbf{p}_{n}-\frac{1}{2} \Delta t \mathbf{V}^{\prime}\left(\mathbf{q}_{n}\right) \\
\mathbf{q}_{n+1} &=\mathbf{q}_{n}+\Delta t \mathbf{K}^{\prime}\left(\mathbf{p}_{n+1 / 2}\right) \\
\mathbf{p}_{n+1} &=\mathbf{p}_{n+1 / 2}-\frac{1}{2} \Delta t \mathbf{V}^{\prime}\left(\mathbf{q}_{n+1}\right)
\end{aligned},
$$
where the subscript denotes the time-step index. For the Hamiltonian system, this algorithm is as efficient as Euler’s method yet more accurate. In addition, it uses $L_2$ loss between the estimated trajectory and the observed trajectory, whereas in HNN the loss is computed via time derivatives. In other words, SRNN learns towards the integration of the Hamiltonian system, while HNN learns the differentiated one. 

Variational Integrator Network(VIN)~\cite{saemundsson2020variational} also applies Velocity Verlet integrator (similar with leapfrog method) to HNN. In addition, VIN introduces Lie group variational integrators (LGVIs), that automatically evolve on a specified Lie group. The key idea is to approximate the change in position over integration steps using Lie group elements. For example, the evolution of the angle of a pendulum in 2D space can be encoded in Lie group $SO(2)$. Recent work \cite{celledoni2022learning} studies one of the Lie group methods, the Runge–Kutta–Munthe–Kaas (RKMK) method on the constrained Hamiltonian system.  

Besides computational issues, symplectic integration is also helpful in improving model efficiency. Neural Interacting Hamiltonian (NIH) \cite{cai2021neural} shows that HNN prediction accuracy and efficiency can be enhanced, if the Hamiltonian can be decomposed as an analytically solvable part and a residual part, and only the small residual part is approximated via a neural network. Such decomposation is: $
\mathcal{H}=\mathcal{H}_{\text {kepler }}+\mathcal{H}_{\text {inter }},$
where $\mathcal{H}_{\text {kepler }}$ denotes the Kepler motion of the bodies with respect to the center of mass, and $\mathcal{H}_{\text {inter }}$ represents the perturbation among the minor bodies. The former can be solved analytically, but the latter has to be solved numerically, i.e. approximated by a network $\mathcal{H}_{\text {inter, } \theta}$. The optimization method is Wisdom-Holman (WH) integrator, a symplectic mapping that employs a drift-kick-drift strategy, where drift step solves Kepler’s equation analytically, and kick step corrects the residual part $\mathcal{H}_{\text {inter }}$.

\subsection{Generalized Input Form}
Depending on the application scenarios, the input form of the Hamiltonian system varies from the canonical coordinate vectors to generalized coordinate vectors, or even the pixel input and graph input, where the hidden states are assumed to satisfy hamiltonian dynamics.

To handle situations when training data in the position-momentum coordinate is unavailable, NSF~\cite{chen2021neural} uses symplectic-2 form, a coordinate-free form of Hamiltonian equations (see Sec.~\ref{sec:general-system}). Symplectic ODE-Net (SymODEN)\cite{zhong2019symplectic} accommodates non-Euclidean coordinates such as angular coordinate, and such angle data is  obtained in the embedded form, i.e., $(\cos q, \sin q)$ instead of the coordinate $(q)$ itself. Hamilton Generative Model(HGN) \cite{toth2019hamiltonian} proposes a Variational Autoencoder(VAE) to accommodate high-dimensional observations (such
as images), and assumes hidden states are governed by Hamiltonian system.

For graph data, Hamiltonian ODE Graph Network (HOGN)~\cite{sanchez2019hamiltonian} combines HNN with graph neural network by the following formulation:
$$
\begin{aligned}
\mathcal{H}_{\mathrm{GN}}(\mathbf{q}, \mathbf{p}) &=\mathrm{GN}_{\mathbf{u}}(\mathbf{q}, \mathbf{p}, \mathbf{c} ; \phi) \\
f_{\dot{\mathbf{q}}, \dot{\mathbf{p}}}^{\mathrm{HOGN}}(\mathbf{q}, \mathbf{p}) & \equiv\left(\frac{\partial \mathcal{H}_{\mathrm{GN}}}{\partial \mathbf{p}},-\frac{\partial \mathcal{H}_{\mathrm{GN}}}{\partial \mathbf{q}}\right)=(\dot{\mathbf{q}}, \dot{\mathbf{p}}) \\
(\mathbf{q}, \mathbf{p})_{n+1} & =\mathrm{RK}\left(\Delta t,(\mathbf{q}, \mathbf{p})_{n}, f_{\dot{\mathbf{q}}, \mathbf{p}}^{\mathrm{HOGN}}\right)
\end{aligned},
$$
where $GN$ denotes a graph network and $RK$ is Runge-Kutta integrator (can be replaced with symplectic integrators). The experiment shows that Hamiltonian inductive bias effectively improves the accuracy and generalization ability of graph neural networks. Similarly, Molecular Hamiltonian Network (HamNet) \cite{li2020conformation} uses a graph neural network to learn the Hamiltonian dynamics of implicit positions and momentum of atoms in a molecule interact. In the HamNet, the chemical and physics prior are explicitly encoded in the expression of $\mathcal{H}$, and also Rayleigh’s dissipation function $\mathcal{D}$, thus it models a generalized Hamiltonian system (Sec.~\ref{sec:general-system}).  

For image sequence or video input, the HNN paper takes the Autoencoder approach, and VIN~\cite{saemundsson2020variational} extends to VAE. \cite{khan2021hamiltonian} further proposes to explicitly disentangle the hidden state into motion and content, while the motion is Hamiltonian dynamics, the content denotes static features like colors and shapes.

\subsection{Extended Problem Settings}
\textbf{Constraints}: Constrained HNN (CHNN)~\cite{finzi2020simplifying} aims at learning constrained Hamiltonian systems. Typically the constraints in physical systems are enforced by generalized coordinates,  e.g. angular, distance, etc. However, CHNN proposes to embed the system into Cartesian coordinates and enforcing the constraints explicitly with Lagrange multipliers, since such configuration dramatically simplifies the learning
problem. \cite{celledoni2022learning} studies Hamiltonian systems that are holonomically constrained on some configuration manifold $\mathcal{Q}=$ $\left\{q \in \mathbb{R}^{n}: g(q)=0\right\}$ embedded in $\mathbb{R}^{n}$. By using embedding property, constrained
multi-body systems can be modeled by means of projection operators. 

\textbf{Meta-Learning}: \cite{lee2020identifying}  aims to train a model well
generalized on new systems governed by the same physical
law but with unperceived physical parameters. They formulate  identifying the shared representation of unknown Hamiltonian systems as a meta-learning problem, and solve the problem using Model-Agnostic Meta Learning (MAML).

\textbf{Control}: Symplectic ODE-Net (SymODEN)\cite{zhong2019symplectic} proposes to add an external control term to the standard Hamiltonian dynamics. With the learned dynamics, SymODEN are able to manipulate controllers to control the system to track a reference configuration. 

\textbf{Flow Model}: HGN \cite{toth2019hamiltonian} proposes a simple modification of HNN that changes it to Neural Hamiltonian Flow (NHF) model. It is especially useful, since the two requirements for normalising flows i.e. invertiblity and volume preserving
are exactly the two basic properties of Hamiltonian dynamics, which can be shown by computing the
determinant of the Jacobian of the infinitesimal transformation induced by the Hamiltonian system:
\begin{eqnarray*}
&&\operatorname{det}\left[\mathbb{I}+d t\left(\begin{array}{cc}
\frac{\partial^{2} \mathcal{H}}{\partial q_{i} \partial p_{j}} & -\frac{\partial^{2} \mathcal{H}}{\partial q_{i} \partial q_{j}} \\
\frac{\partial^{2} \mathcal{H}}{\partial p_{i} \partial p_{j}} & -\frac{\partial^{2} \mathcal{H}}{\partial p_{i} \partial q_{j}}
\end{array}\right)\right] \\
&&= 1+d t \operatorname{Tr}\left(\begin{array}{cc}
\frac{\partial^{2} \mathcal{H}}{\partial q_{i} \partial p_{j}} & -\frac{\partial^{2} \mathcal{H}}{\partial q_{i} \partial q_{j}} \\
\frac{\partial^{2} \mathcal{H}}{\partial p_{i} \partial p_{j}} & -\frac{\partial^{2} \mathcal{H}}{\partial p_{i} \partial q_{j}}
\end{array}\right)+O\left(d t^{2}\right)\\
&&=1+O\left(d t^{2}\right)
\end{eqnarray*}

\section{Limitations and Future Directions}
Priors and assumptions limit HNN's applicability. Intuitively, there is no free lunch. The original neural networks have universal approximation ability in theory, while HNN is only suitable for energy-conserving dynamics. DHNN extends HNN to non-conserving situations, but with additional assumptions on the energy dissipation system. Other variants of HNN are also subject to various forms of assumption. For example, as discussed in Sec.~\ref{sec:symplectic-integration}, SRNN with leapfrog integrator assumes the Hamiltonian system separable. 

Hamiltonian system can be \emph{chaotic}, characterized by a sensitive dependence on initial conditions, which means the same system with slightly different initial conditions will diverge exponentially in time. Well-known examples include the three-body system, Hénon-Heiles system, and zero-sum game, and are also common in weather, fluid, and celestial systems. Experiments in \cite{chen2019symplectic,choudhary2020physics} show that vanilla HNN is not able to accurately recover three-body system and Hénon-Heiles system, while symplectic integrator can effectively improve the performance in the chaotic systems \cite{dipietro2020sparse}. However, as an intrinsic problem of the Hamiltonian system, chaos represent the differential system can be extremely ill-conditioned, which is intuitively not suitable for learning and is often not expected in real applications, hindering the applications of such models.

Technically speaking, the existing Hamiltonian-based learning frameworks are plagued with other issues which require future works. First, Hamiltonian systems are restricted to the time translation invariance and the energy conservation law, which does not hold in some physical cases like inelastic collisions. Second, on solving the underlying differential equations, Hamiltonian-based learning frameworks need extra efforts of integration to derive the solution of the ODE system, which is inferior to works like Neural Fourier Operators in test-time computational efficiency. Third, in the training phase, classical models incorporate partial derivatives of the Hamiltonian in its loss either explicitly (pointwise form) or implicitly (quadrature form), which adds up to the computational overhead of the backward process, where the Jacobian of these terms are calculated.

We have summarized the popular models to show the comparison of adopted methodologies in Table.~\ref{tab:1}.

\section{Open-source codes and Datasets}
We give an open-source list of HNN and its variants. The authors of HNN provide a well-annotated source code\footnote{\url{https://github.com/greydanus/hamiltonian-nn.git}}. For the symplectic integrator, SRNN is a nice introductory example with plenty of ablation experiments\footnote{\url{ https://github.com/zhengdao-chen/SRNN.git}}. For generalized Hamiltionian system, GHNN provides detailed tutorials with visualizations\footnote{\url{https://github.com/coursekevin/weakformghnn.git}}.

We also list a few representative datasets in this emerging area: 1) OpenAI Gym’s \emph{Pendulum-v0}~\cite{brockman2016openai}, an environment that produces trajectories of $400 \times 400 \times 3$ RGB pixel observations of a pendulum arm; 2) Ideal mass-spring and damped-spring, the simplest physics setting for Hamiltonian and generalized Hamiltonian systems; 3) Three-Body system, a typical chaotic dynamic system.

\section{Conclusion and Outlook}
The learning and prediction of Hamiltonian system is a fundamental and promising direction that symbolizes the basic forward and inverse problems in differential equations. The recent advances of generalizing formulation of Hamilton's system and symplecticity-enforced neural architectures in this line have shown great advantages in prediction accuracy and sample efficiency etc. 

With emerging new techniques, the idea of Hamiltonian-based learning will hopefully lead to more advanced deep learning architectures incorporating physical information. We proactively anticipate the applications of Hamiltonian-based learning to broader domains beyond physics.

\bibliographystyle{named}
\bibliography{reference}

\begin{thebibliography}{}

\bibitem[\protect\citeauthoryear{Amos \bgroup \em et al.\egroup
  }{2017}]{amos2017input}
Brandon Amos, Lei Xu, and J~Zico Kolter.
\newblock Input convex neural networks.
\newblock In {\em ICML}, 2017.

\bibitem[\protect\citeauthoryear{Bloch \bgroup \em et al.\egroup
  }{2001}]{bloch2001controlled}
Anthony~M Bloch, Naomi~Ehrich Leonard, and Jerrold~E Marsden.
\newblock Controlled lagrangians and the stabilization of euler--poincar{\'e}
  mechanical systems.
\newblock {\em International Journal of Robust and Nonlinear Control:
  IFAC-Affiliated Journal}, 11(3):191--214, 2001.

\bibitem[\protect\citeauthoryear{Brockman \bgroup \em et al.\egroup
  }{2016}]{brockman2016openai}
Greg Brockman, Vicki Cheung, Ludwig Pettersson, Jonas Schneider, John Schulman,
  Jie Tang, and Wojciech Zaremba.
\newblock Openai gym.
\newblock {\em arXiv preprint arXiv:1606.01540}, 2016.

\bibitem[\protect\citeauthoryear{Cai \bgroup \em et al.\egroup
  }{2021}]{cai2021neural}
Maxwell~X Cai, Simon~Portegies Zwart, and Damian Podareanu.
\newblock Neural symplectic integrator with hamiltonian inductive bias for the
  gravitational $ n $-body problem.
\newblock {\em arXiv preprint arXiv:2111.15631}, 2021.

\bibitem[\protect\citeauthoryear{Celledoni \bgroup \em et al.\egroup
  }{2022}]{celledoni2022learning}
Elena Celledoni, Andrea Leone, Davide Murari, and Brynjulf Owren.
\newblock Learning hamiltonians of constrained mechanical systems.
\newblock {\em arXiv preprint arXiv:2201.13254}, 2022.

\bibitem[\protect\citeauthoryear{Chen and Tao}{2021}]{chen2021data}
Renyi Chen and Molei Tao.
\newblock Data-driven prediction of general hamiltonian dynamics via learning
  exactly-symplectic maps.
\newblock In {\em ICML}, pages 1717--1727. PMLR, 2021.

\bibitem[\protect\citeauthoryear{Chen \bgroup \em et al.\egroup
  }{2018}]{chen2018neural}
Ricky~TQ Chen, Yulia Rubanova, Jesse Bettencourt, and David~K Duvenaud.
\newblock Neural ordinary differential equations.
\newblock {\em NeurIPS}, 31, 2018.

\bibitem[\protect\citeauthoryear{Chen \bgroup \em et al.\egroup
  }{2019}]{chen2019symplectic}
Zhengdao Chen, Jianyu Zhang, Martin Arjovsky, and L{\'e}on Bottou.
\newblock Symplectic recurrent neural networks.
\newblock {\em arXiv preprint arXiv:1909.13334}, 2019.

\bibitem[\protect\citeauthoryear{Chen \bgroup \em et al.\egroup
  }{2021}]{chen2021neural}
Yuhan Chen, Takashi Matsubara, and Takaharu Yaguchi.
\newblock Neural symplectic form: Learning hamiltonian equations on general
  coordinate systems.
\newblock {\em NeurIPS}, 2021.

\bibitem[\protect\citeauthoryear{Choudhary \bgroup \em et al.\egroup
  }{2020}]{choudhary2020physics}
Anshul Choudhary, John~F Lindner, Elliott~G Holliday, Scott~T Miller, Sudeshna
  Sinha, and William~L Ditto.
\newblock Physics-enhanced neural networks learn order and chaos.
\newblock {\em Physical Review E}, 101(6):062207, 2020.

\bibitem[\protect\citeauthoryear{Course \bgroup \em et al.\egroup
  }{2020}]{course2020weak}
Kevin Course, Trefor Evans, and Prasanth Nair.
\newblock Weak form generalized hamiltonian learning.
\newblock {\em NeurIPS}, 2020.

\bibitem[\protect\citeauthoryear{DiPietro \bgroup \em et al.\egroup
  }{2020}]{dipietro2020sparse}
Daniel~M DiPietro, Shiying Xiong, and Bo~Zhu.
\newblock Sparse symplectically integrated neural networks.
\newblock {\em arXiv preprint arXiv:2006.12972}, 2020.

\bibitem[\protect\citeauthoryear{Dutt \bgroup \em et al.\egroup
  }{2021}]{dutt2021active}
Arkopal Dutt, Edwin Pednault, Chai~Wah Wu, Sarah Sheldon, John Smolin, Lev
  Bishop, and Isaac~L Chuang.
\newblock Active learning of quantum system hamiltonians yields query
  advantage.
\newblock {\em arXiv preprint arXiv:2112.14553}, 2021.

\bibitem[\protect\citeauthoryear{Finzi \bgroup \em et al.\egroup
  }{2020}]{finzi2020simplifying}
Marc Finzi, Ke~Alexander Wang, and Andrew~G Wilson.
\newblock Simplifying hamiltonian and lagrangian neural networks via explicit
  constraints.
\newblock {\em NeurIPS}, 2020.

\bibitem[\protect\citeauthoryear{Greydanus and
  Sosanya}{2022}]{greydanus2022dissipative}
Sam Greydanus and Andrew Sosanya.
\newblock Dissipative hamiltonian neural networks: Learning dissipative and
  conservative dynamics separately.
\newblock {\em arXiv preprint arXiv:2201.10085}, 2022.

\bibitem[\protect\citeauthoryear{Greydanus \bgroup \em et al.\egroup
  }{2019}]{greydanus2019hamiltonian}
Samuel Greydanus, Misko Dzamba, and Jason Yosinski.
\newblock Hamiltonian neural networks.
\newblock {\em NeurIPS}, 2019.

\bibitem[\protect\citeauthoryear{Haier \bgroup \em et al.\egroup
  }{2006}]{haier2006geometric}
Ernst Haier, Christian Lubich, and Gerhard Wanner.
\newblock {\em Geometric Numerical integration: structure-preserving algorithms
  for ordinary differential equations}.
\newblock Springer, 2006.

\bibitem[\protect\citeauthoryear{Han \bgroup \em et al.\egroup
  }{2018}]{han2018solving}
Jiequn Han, Arnulf Jentzen, and E~Weinan.
\newblock Solving high-dimensional partial differential equations using deep
  learning.
\newblock {\em Proceedings of the National Academy of Sciences},
  115(34):8505--8510, 2018.

\bibitem[\protect\citeauthoryear{Helmholtz}{1858}]{helmholtz1858integrale}
H~von Helmholtz.
\newblock {\"U}ber integrale der hydrodynamischen gleichungen, welche den
  wirbelbewegungen entsprechen.
\newblock 1858.

\bibitem[\protect\citeauthoryear{Howse \bgroup \em et al.\egroup
  }{1995}]{howse1995gradient}
James Howse, Chaouki Abdallah, and Gregory Heileman.
\newblock Gradient and hamiltonian dynamics applied to learning in neural
  networks.
\newblock {\em NeurIPS}, 1995.

\bibitem[\protect\citeauthoryear{Khan and Storkey}{2021}]{khan2021hamiltonian}
Asif Khan and Amos Storkey.
\newblock Hamiltonian prior to disentangle content and motion in image
  sequences.
\newblock {\em arXiv preprint arXiv:2112.01641}, 2021.

\bibitem[\protect\citeauthoryear{Kovachki \bgroup \em et al.\egroup
  }{2021}]{kovachki2021neural}
Nikola Kovachki, Zongyi Li, Burigede Liu, Kamyar Azizzadenesheli, Kaushik
  Bhattacharya, Andrew Stuart, and Anima Anandkumar.
\newblock Neural operator: Learning maps between function spaces.
\newblock {\em arXiv preprint arXiv:2108.08481}, 2021.

\bibitem[\protect\citeauthoryear{Lee \bgroup \em et al.\egroup
  }{2020}]{lee2020identifying}
Seungjun Lee, Haesang Yang, and Woojae Seong.
\newblock Identifying physical law of hamiltonian systems via meta-learning.
\newblock In {\em International Conference on Learning Representations}, 2020.

\bibitem[\protect\citeauthoryear{Lee \bgroup \em et al.\egroup
  }{2021}]{lee2021machine}
Kookjin Lee, Nathaniel Trask, and Panos Stinis.
\newblock Machine learning structure preserving brackets for forecasting
  irreversible processes.
\newblock {\em NeurIPS}, 2021.

\bibitem[\protect\citeauthoryear{Leimkuhler and
  Reich}{2004}]{leimkuhler2004simulating}
Benedict Leimkuhler and Sebastian Reich.
\newblock {\em Simulating hamiltonian dynamics}.
\newblock Number~14. Cambridge university press, 2004.

\bibitem[\protect\citeauthoryear{Letcher \bgroup \em et al.\egroup
  }{2019}]{letcher2019differentiable}
Alistair Letcher, David Balduzzi, S{\'e}bastien Racaniere, James Martens, Jakob
  Foerster, Karl Tuyls, and Thore Graepel.
\newblock Differentiable game mechanics.
\newblock {\em The Journal of Machine Learning Research}, 20(1):3032--3071,
  2019.

\bibitem[\protect\citeauthoryear{Li \bgroup \em et al.\egroup
  }{2020a}]{li2020conformation}
Ziyao Li, Shuwen Yang, Guojie Song, and Lingsheng Cai.
\newblock Conformation-guided molecular representation with hamiltonian neural
  networks.
\newblock In {\em ICLR}, 2020.

\bibitem[\protect\citeauthoryear{Li \bgroup \em et al.\egroup
  }{2020b}]{li2020fourier}
Zongyi Li, Nikola Kovachki, Kamyar Azizzadenesheli, Burigede Liu, Kaushik
  Bhattacharya, Andrew Stuart, and Anima Anandkumar.
\newblock Fourier neural operator for parametric partial differential
  equations.
\newblock {\em arXiv preprint arXiv:2010.08895}, 2020.

\bibitem[\protect\citeauthoryear{Lutter \bgroup \em et al.\egroup
  }{2018}]{lutter2018deep}
Michael Lutter, Christian Ritter, and Jan Peters.
\newblock Deep lagrangian networks: Using physics as model prior for deep
  learning.
\newblock In {\em ICLR}, 2018.

\bibitem[\protect\citeauthoryear{Matsubara \bgroup \em et al.\egroup
  }{2021}]{matsubara2021symplectic}
Takashi Matsubara, Yuto Miyatake, and Takaharu Yaguchi.
\newblock Symplectic adjoint method for exact gradient of neural ode with
  minimal memory.
\newblock {\em NeurIPS}, 2021.

\bibitem[\protect\citeauthoryear{Ortega \bgroup \em et al.\egroup
  }{2002}]{ortega2002interconnection}
Romeo Ortega, Arjan Van Der~Schaft, Bernhard Maschke, and Gerardo Escobar.
\newblock Interconnection and damping assignment passivity-based control of
  port-controlled hamiltonian systems.
\newblock {\em Automatica}, 38(4):585--596, 2002.

\bibitem[\protect\citeauthoryear{Poincar{\'e}}{1899}]{poincare1899methodes}
Henri Poincar{\'e}.
\newblock {\em Les m{\'e}thodes nouvelles de la m{\'e}canique c{\'e}leste},
  volume~3.
\newblock Gauthier-Villars, 1899.

\bibitem[\protect\citeauthoryear{Polydoros and
  Nalpantidis}{2017}]{polydoros2017survey}
Athanasios~S Polydoros and Lazaros Nalpantidis.
\newblock Survey of model-based reinforcement learning: Applications on
  robotics.
\newblock {\em Journal of Intelligent \& Robotic Systems}, 86(2):153--173,
  2017.

\bibitem[\protect\citeauthoryear{Raissi \bgroup \em et al.\egroup
  }{2019}]{raissi2019physics}
Maziar Raissi, Paris Perdikaris, and George~E Karniadakis.
\newblock Physics-informed neural networks: A deep learning framework for
  solving forward and inverse problems involving nonlinear partial differential
  equations.
\newblock {\em Journal of Computational physics}, 378:686--707, 2019.

\bibitem[\protect\citeauthoryear{Rupp \bgroup \em et al.\egroup
  }{2012}]{rupp2012fast}
Matthias Rupp, Alexandre Tkatchenko, Klaus-Robert M{\"u}ller, and O~Anatole
  Von~Lilienfeld.
\newblock Fast and accurate modeling of molecular atomization energies with
  machine learning.
\newblock {\em Physical review letters}, 108(5):058301, 2012.

\bibitem[\protect\citeauthoryear{Saemundsson \bgroup \em et al.\egroup
  }{2020}]{saemundsson2020variational}
Steindor Saemundsson, Alexander Terenin, Katja Hofmann, and Marc Deisenroth.
\newblock Variational integrator networks for physically structured embeddings.
\newblock In {\em AISTATS}, 2020.

\bibitem[\protect\citeauthoryear{Sanchez-Gonzalez \bgroup \em et al.\egroup
  }{2019}]{sanchez2019hamiltonian}
Alvaro Sanchez-Gonzalez, Victor Bapst, Kyle Cranmer, and Peter Battaglia.
\newblock Hamiltonian graph networks with ode integrators.
\newblock {\em arXiv preprint arXiv:1909.12790}, 2019.

\bibitem[\protect\citeauthoryear{Seung \bgroup \em et al.\egroup
  }{1997}]{seung1997minimax}
H~Sebastian Seung, Tom Richardson, J~Lagarias, and John~J Hopfield.
\newblock Minimax and hamiltonian dynamics of excitatory-inhibitory networks.
\newblock {\em NeurIPS}, 1997.

\bibitem[\protect\citeauthoryear{Toth \bgroup \em et al.\egroup
  }{2019}]{toth2019hamiltonian}
Peter Toth, Danilo~J Rezende, Andrew Jaegle, S{\'e}bastien Racani{\`e}re,
  Aleksandar Botev, and Irina Higgins.
\newblock Hamiltonian generative networks.
\newblock In {\em International Conference on Learning Representations}, 2019.

\bibitem[\protect\citeauthoryear{Valenti \bgroup \em et al.\egroup
  }{2019}]{valenti2019hamiltonian}
Agnes Valenti, Evert van Nieuwenburg, Sebastian Huber, and Eliska Greplova.
\newblock Hamiltonian learning for quantum error correction.
\newblock {\em Physical Review Research}, 1(3):033092, 2019.

\bibitem[\protect\citeauthoryear{Wang and Yu}{2021}]{wang2021physics}
Rui Wang and Rose Yu.
\newblock Physics-guided deep learning for dynamical systems: A survey.
\newblock {\em arXiv preprint arXiv:2107.01272}, 2021.

\bibitem[\protect\citeauthoryear{Weinan and Yu}{2018}]{weinan2018deep}
E~Weinan and Bing Yu.
\newblock The deep ritz method: A deep learning-based numerical algorithm for
  solving variational problems.
\newblock {\em Communications in Mathematics and Statistics}, 1(6):1--12, 2018.

\bibitem[\protect\citeauthoryear{Willard \bgroup \em et al.\egroup
  }{2020}]{willard2020integrating}
Jared Willard, Xiaowei Jia, Shaoming Xu, Michael Steinbach, and Vipin Kumar.
\newblock Integrating physics-based modeling with machine learning: A survey.
\newblock {\em arXiv preprint arXiv:2003.04919}, 1(1):1--34, 2020.

\bibitem[\protect\citeauthoryear{Yang \bgroup \em et al.\egroup
  }{2017}]{yang2017hamiltonian}
Yongliang Yang, Donald Wunsch, and Yixin Yin.
\newblock Hamiltonian-driven adaptive dynamic programming for continuous
  nonlinear dynamical systems.
\newblock {\em IEEE transactions on neural networks and learning systems},
  28(8):1929--1940, 2017.

\bibitem[\protect\citeauthoryear{Zhong \bgroup \em et al.\egroup
  }{2019}]{zhong2019symplectic}
Yaofeng~Desmond Zhong, Biswadip Dey, and Amit Chakraborty.
\newblock Symplectic ode-net: Learning hamiltonian dynamics with control.
\newblock In {\em International Conference on Learning Representations}, 2019.

\bibitem[\protect\citeauthoryear{Zhong \bgroup \em et al.\egroup
  }{2021}]{zhong2021benchmarking}
Yaofeng~Desmond Zhong, Biswadip Dey, and Amit Chakraborty.
\newblock Benchmarking energy-conserving neural networks for learning dynamics
  from data.
\newblock In {\em Learning for Dynamics and Control}, 2021.

\end{thebibliography}

\end{document}